# Identifying Confusion Trends in Concept-based XAI for Multi-Label Classification

Haadia Amjad
Chair of Fundamentals of Electrical Engineering
TUD Dresden University of Technology
Dresden, Germany
e-mail: haadia.amjad@tu-dresden.de

Kilian Göller
Chair of Fundamentals of Electrical Engineering
TUD Dresden University of Technology
Dresden, Germany
e-mail: kilian.goeller@tu-dresden.de

Steffen Seitz
Chair of Fundamentals of Electrical Engineering
TUD Dresden University of Technology
Dresden, Germany
e-mail: steffen.seitz@tu-dresden.de

Carsten Knoll
Chair of Fundamentals of Electrical Engineering
TUD Dresden University of Technology
Dresden, Germany
e-mail: carsten.knoll@tu-dresden.de

Ronald Tetzlaff
Chair of Fundamentals of Electrical Engineering
TUD Dresden University of Technology
Dresden, Germany
e-mail: ronald.tetzlaff@tu-dresden.de

***Abstract*—Deep Neural Networks (DNNs) deployed in high-risk domains, such as healthcare and autonomous driving, must be not only accurate but also understandable to ensure user trust. In real-world computer vision tasks, these models often operate on complex images containing background noise and are heavily annotated. To make such models explainable, Concept-based Explainable AI (CXAI) methods need to be assessed for their applicability and problem-solving capacity. In this work, we explore CXAI use cases in multi-label classification by training two DNNs, VGG16 and ResNet50, on the 20 most annotated labels in the MS-COCO dataset (Microsoft Common Objects in Context). We apply two CXAI methods, CRP (Concept Relevance Propagation) and CRAFT (Concept Recursive Activation FacTorization), to generate concept-level explanations and investigate the overall evaluations. Our analysis reveals three key findings: (1) CXAI highlights learning weaknesses in DNNs, (2) higher concept distinctiveness reduces label and concept confusion, and (3) environmental concepts expose dataset-induced biases. Our results demonstrate the potential of CXAI to enhance the understanding of model generalizability and to diagnose bias instigated by the dataset.**

## I. INTRODUCTION

Deep Neural Network (DNN) [1] performance is crucial for their adoption in real-world applications. However, understanding their decisions is also important, especially in high-risk domains like autonomous driving and medical diagnosis. Real-world datasets often vary in resolution and object size, with complex scenes including small, clustered, or overlapping objects. Multi-label datasets, where images have multiple annotations, frequently suffer from class imbalance. This can lead to confusion (i.e., errors made in predicting the correct class/data points) between labels and wrong associations. Even high-performing models that exhibit confusion need deeper analysis. Explainable AI (XAI) methods are useful in revealing these learning patterns [2].

XAI provides interpretability for black-box models [2]. Concept-based XAI (CXAI) identifies semantically meaningful features relevant to a class [3], unlike saliency maps, which are harder to interpret in complex scenes [4]. Concepts reflect how a DNN internally represents a class [5]. However, DNNs may learn unintended associations, concept bias or spurious correlations, where background elements influence classification (e.g., associating "fingers" with a pen) [6]. We refer to non-target concepts produced by such bias as "environmental concepts."

CXAI methods often visualize activation maps or focused image regions [7]. These show both target and environmental concepts. Determining whether an environmental concept is valid requires further analysis. Its presence may reflect dataset bias or mislearning.

In this work, we train two state-of-the-art DNNs, ResNet50 and VGG-16, on the 20 most annotated MS-COCO labels [8]. Using two model checkpoints per architecture, one well-performing and one poor, we evaluate their predictions using CXAI methods: CRP [9] and CRAFT [10]. These methods produce focused region visualizations and scores that determine a concept's contribution to the overall learning (concept importance) or target label learning (concept relevance) of the DNN model. We compare results using concept error and distinctiveness (see Section III, D) to study confusion trends across models.

***The main contributions of this paper can be summarized as follows:***

- We demonstrate that CXAI methods can reveal learning weaknesses in deep neural networks.
- We find that greater concept distinctiveness is associated with reduced confusion in label predictions and concept attributions.
- We show that environmental concepts can expose dataset-induced biases in model learning and interpretation.

The remainder of this paper is organized as follows: Section II reviews related studies. Section III describes the experimental setup, including the dataset, DNN models, CXAI methods, and key terminology. Section IV presents the results structured around our three main contributions. Finally, Section V concludes the paper and discusses directions for future work.

## II. Related Work

Various CXAI methods are available for use today, and it is a growing research field. Lee et al. [11] detail the current state of CXAI methods. Their study identifies three main directions for future research: the choice of concepts to explain, the selection of concept representation, and methods to control concepts.

Some studies focus on using concepts to detect potential biases in DNN models. Their evaluation emphasizes the relationship between different concepts and classes and aims to expose potential biases in the learning of the DNN. Singh et al. [12] study model biases in both, the model learning process and the model's semantic understanding (concept biases), by evaluating the DNN model's ability to recognize a class in the presence and absence of the established context (via learning) for a multi-label classification task.

With newer emerging methods in the realm of CXAI, the desire to fully understand how they can be effectively used with AI systems increases. The dataset, for example, is an important factor contributing to the meaningfulness of the explainability method. The evaluation of CXAI by Ramaswamy et al. [13] addresses important considerations for CXAI methods that influence their effective usage. They emphasize that the impact of the choice of the dataset, even with slight variations in the dataset options, changes the model decision and the explanation provided by a CXAI method.

To study the relationship between confusion and concept-based explanations, we select two CXAI methods to answer the "where" (..*the important information is*) and "what" (..*is the important information*) questions. CRP, proposed by Achtibat et al. [9], is based on the Layer-wise Relevance Propagation (LRP) method [14]. CRP addresses "what" and "where" explanations by exploiting concepts in hidden layers of a DNN model and locating them in the input data. It assesses the contribution of each concept for a target class; in other words, it introduces concept relevance. CRP utilizes relevance maximization to tune its visualization, which depicts a series of focused concepts. CRAFT is another "what" and "where" method proposed by Fel et al. [10] based on the Grad-CAM method [15]. They utilize Sobol indices to estimate the importance of concepts that have been identified using Non-Negative Matrix Factorization (NMF) recursively, generating sub-concepts (concepts of smaller, more focused areas in the image).

Existing research has advanced CXAI by defining concepts and applying them to detect biases and assess dataset effects. Building on this foundation, our work investigates how confusion interacts with concept-based explanations through the lens of CRP and CRAFT.

## III. Experimental Setup

Just as with any other explainable AI pipeline, our experimentation contains the training of DNNs model and its evaluations and the usage of an XAI method and its evaluation, illustrated in Figure 1. This section contains details of our workflow.

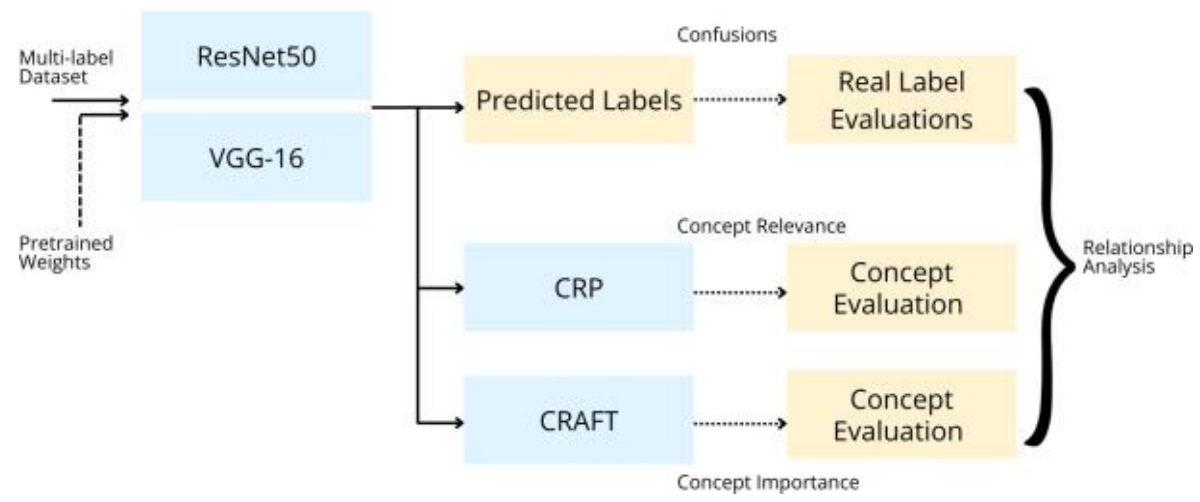

Figure 1. Schematic diagram of our experimental setup.

### A. Dataset

MS-COCO [8] is a large-scale dataset widely used for computer vision tasks such as object detection, captioning, segmentation and classification. The 2017 object detection subset includes 80 "things" classes, objects with clear boundaries, across 118,000 images. As test labels are unavailable, we split the training set 90/10, resulting in 106,200 training and 11,800 test images. For our experiments, we focus on the 20 most frequently annotated labels in the training set to ensure sufficient data per class and meaningful inter label relationships.

### B. DNN Models

ResNet50 [16] is a popular image classification model due to its residual learning feature, which mitigates information loss. It balances accuracy and efficiency well, and its ImageNet-pretrained weights are widely used [17].

VGG-16 [17], known for its simple and uniform structure of stacked convolutional and fully connected layers, is often used as a baseline for deep learning applications. Despite its larger parameter count, it performs well on classification tasks and is easy to implement.

These two models are chosen as they are widely used, and many XAI methods have been proven to work with them. Some of the latest models require large adaptations of XAI methods to be made [18]. Our study focuses on base-level use cases, to be adaptable across different domains; hence, we train ResNet50 and VGG-16 models, pretrained on ImageNetV2 [19], using PyTorch for 350 epochs, saving all checkpoints. For each model, we select two “scenarios” from the saved checkpoints:

- **Scenario 1 (well-performing model):**
  - **ResNet50:** Accuracy: 82.85%, Recall: 85.50, Precision: 58.84, F1 Score: 60.84
  - **VGG-16:** Accuracy: 84.26%, Recall: 86.91, Precision: 59.74, F1 Score: 58.84
- **Scenario 2 (poor-performing model):**
  - **ResNet50:** Accuracy: 58.24%, Recall: 77.04, Precision: 53.82, F1 Score: 42.92
  - **VGG-16:** Accuracy: 52.85%, Recall: 74.50, Precision: 53.62, F1 Score: 46.12

These scenarios are created to have two different sets of performance metrics against which to evaluate explainability. We evaluate models using accuracy, recall, precision, F1 score, and confusion matrices tailored for multi-label tasks. Specifically, we use the multi-label confusion tensor by Krstinić et al. [20], which accounts for label imbalance—well-suited for the MS-COCO dataset.

We also compute Mutual Information (MI) and Jaccard Similarity Coefficient (JSC) between labels. We use these metrics to understand which target labels are more likely to share information or similarities with which predicted labels.

### C. CXAI Methods

We investigate the effect of confusion on two CXAI methods, CRAFT and CRP, across all four model scenarios.

- **CRAFT** outputs concept *importance*, representing the overall contribution of each concept to the model’s learning process.
- **CRP** provides concept *relevance*, indicating the contribution of a concept to specific target classes.

While both methods offer different perspectives, we do not compare them directly or suggest one is superior. Instead, we use their outputs to explore how label confusion is reflected in learned concepts.

We compute concept distinctiveness [21] and concept error [22] for both methods. Concept error is evaluated against a subjective ground truth (detailed in the next section). Additionally, we adapt mutual information to measure shared information between concepts and compare these findings to our DNN evaluations to support our hypotheses.

### D. Explanation of terms (in brief)

This sub-section briefly explains some terminologies in CXAI and our adaptations.

*1) Concept Distinctiveness:* Concept distinctiveness, defined in Eq. (1), measures how unique a concept is compared to others, with values ranging from 0 to 1. Low distinctiveness suggests overlapping or redundant concepts, which may indicate learning errors [21].

$$D(C_i, C_j) = 1 - \frac{v_{C_i} \cdot v_{C_j}}{|v_{C_i}||v_{C_j}|} \tag{1}$$

Here, $v_{C_i}$ and $v_{C_j}$ are the concept vectors for concepts $C_i$ and $C_j$, respectively. Concept vectors are directions in activation space that capture distinct features [23].

*2) Concept Error:* Concept error captures incorrect or irrelevant concept usage during prediction [22]. To approximate accuracy (in binary classification), we define a rough “ground truth” by selecting only those concepts that belong to the target class, excluding environmental concepts. This approach offers an estimate of model confusion, though a structured human study is recommended for practical validation.

*3) Mutual Information:* Mutual information (MI) quantifies the dependency between two variables. In multi-label classification, it measures how much information one label provides about another. Applied to concepts, MI reflects how much information is shared between two concept vectors, revealing potential dependencies or redundancies in learned features [24].

## IV. Results

In this section, we present our findings based on case studies of different label evaluations. These case studies comprise comparisons of the evaluations described in the previous section.

### A. Confusion in Labels Can Be Understood by Their Explanations

TABLE I. TOP CONFUSION AND MUTUAL INFORMATION SCORES IN SCENARIO 1 OF RESNET50

| Class Name | Top Confusion Class | | | | Top MI Class | | | | Jaccard Similarity |
|---|---|---|---|---|---|---|---|---|---|
| | 1st | Score | 2nd | Score | 1st | MI Score | 2nd | MI Score | |
| person | car | 1148.00 | chair | 1081.70 | handbag | 0.0221 | backpack | 0.0176 | 0.6008 |
| car | truck | 207.17 | bench | 173.16 | truck | 0.0400 | traffic light | 0.0280 | 0.1894 |
| motorcycle | truck | 86.33 | handbag | 85.70 | car | 0.0094 | person | 0.0037 | 0.1474 |
| truck | airplane | 118.35 | car | 117.22 | car | 0.0399 | boat | 0.0020 | 0.0077 |
| boat | Parking meter | 89.70 | car | 76.30 | chair | 0.0682 | fork | 0.0017 | 0.0068 |

TABLE II. TOP CONFUSION AND MUTUAL INFORMATION SCORES IN SCENARIO 2 OF RESNET50

| Class Name | Top Confusion Class | | | | Top MI Class | | | | Jaccard Similarity |
|---|---|---|---|---|---|---|---|---|---|
| | 1st | Score | 2nd | Score | 1st | MI Score | 2nd | MI Score | |
| person | backpack | 1995.20 | bench | 1922.50 | tie | 0.0221 | umbrella | 0.0176 | 0.5470 |
| car | backpack | 340.70 | bench | 334.80 | boat | 0.0399 | stop sign | 0.0280 | 0.1159 |
| motorcycle | backpack | 277.60 | handbag | 273.41 | bicycle | 0.0372 | car | 0.0199 | 0.1289 |
| truck | backpack | 209.07 | bench | 200.09 | motorcycle | 0.0399 | Fire hydrant | 0.0077 | 0.0755 |
| boat | car | 134.04 | bird | 133.66 | fork | 0.0017 | refrigerator | 0.0010 | 0.0440 |

TABLE III. PERCENTAGE OF CO-OCCURRENCE OF TARGET LABEL WITH OTHER LABELS (TOP 20 FREQUENTLY ANNOTATED LABELS)

| Class Name | 1st Class | % | 2nd Class | % |
|---|---|---|---|---|
| person | car | 13.29 | backpack | 7.85 |
| car | person | 69.54 | backpack | 8.43 |
| motorcy -cle | person | 79.55 | car | 39.32 |
| truck | person | 65.15 | car | 59.80 |
| boat | person | 65.69 | car | 8.66 |
| traffic light | car | 61.22 | person | 59.19 |
| bench | person | 73.75 | car | 14.63 |
| bird | person | 24.56 | boat | 7.29 |
| sheep | person | 24.07 | dog | 7.59 |
| backpack | person | 91.06 | car | 18.69 |
| umbrella | person | 86.87 | handbag | 28.81 |
| handbag | person | 90.95 | backpack | 24.62 |
| kite | person | 92.84 | car | 11.54 |
| bottle | person | 53.65 | cup | 34.65 |
| cup | person | 52.76 | dining table | 50.92 |
| bowl | dining table | 47.76 | person | 40.73 |
| banana | person | 41.37 | bowl | 23.05 |
| potted plant | person | 44.07 | chair | 38.61 |
| dining table | person | 49.58 | chair | 43.29 |
| book | dining table | 75.61 | cup | 52.97 |

TABLE IV. CXAI METHOD EVALUATION COMPARED WITH CONFUSION SCORE FOR 'PERSON' LABEL

| Label | Model | CXAI Method | Concept Error | Concept Distinct -iveness | Confus -ion Score |
|---|---|---|---|---|---|
| Person | ResNet Scenario 1 | Craft | 0.20 | 0.76 | 0.09 |
| Person | ResNet Scenario 2 | Craft | 0.38 | 0.48 | 0.26 |
| Person | VGG-16 Scenario 1 | Craft | 0.24 | 0.71 | 0.12 |
| Person | VGG-16 Scenario 2 | Craft | 0.41 | 0.43 | 0.28 |

Label confusion occurs when models struggle to distinguish between classes with overlapping features or co-occurring contexts, often due to ambiguous data, mislabeling, or internal misinterpretation. We hypothesize that CXAI methods, particularly through MI and concept distinctiveness, can reveal whether confusion stems from visual similarity, dataset bias, or how the model encodes relationships between labels.

Tables I and II present confusion and MI scores for three highly confused classes across both ResNet50 scenarios. In scenario 1, *person* is confused with *car* and *chair*, while *car* overlaps with *truck* and *bench*. MI analysis shows that *person* shares high information content with *handbag* and *backpack*, and *car* with *truck* and *traffic light*. These associations indicate that the model is not learning isolated class-specific features, but instead forming dependencies based on recurring visual or contextual co-occurrence. Table III supports this, showing frequent joint appearance of labels such as *person* and *accessories*, or *car* and *truck*, which reinforces these spurious links.

Table IV further highlights the role of CXAI metrics in understanding confusion. In scenario 1, where models perform better, *person* has lower concept error and higher distinctiveness, aligning with reduced confusion. In scenario 2, we observe the opposite: increased concept error, lower distinctiveness, and significantly higher confusion scores. These patterns suggest that when a model lacks distinct conceptual boundaries between classes, it tends to rely more heavily on misleading contextual aspects.

Together, these findings show how CXAI methods help expose the roots of confusion. By combining explanations with performance metrics and co-occurrence statistics, we gain a clearer view of when confusion reflects real-world visual similarity versus when it results from dataset bias or poor internal representations.

### B. *Distinctiveness Reduces Conceptual Confusion*

When a concept is distinct, its features are unique and specific, allowing it to be more accurately defined and recognized. In contrast, concepts derived from confused or overlapping labels tend to be “confused” themselves, as they learn features that are shared across multiple classes rather than those unique to their true class. This issue arises from concept bias, where the model may associate a class with irrelevant features that co-occur with other classes, as shown in Figure 2.

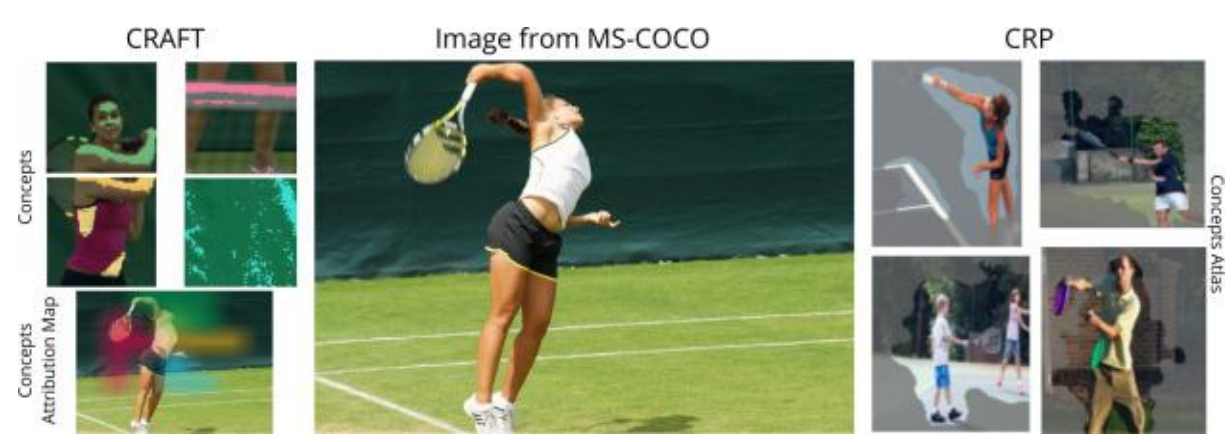

Figure 2. Concepts of class “tennis racket” in scenario 1 of VGG-16. We can see that “person” is heavily present in these explanations.

TABLE V. MUTUAL INFORMATION, CONCEPT DISTINCTIVENESS, AND CONCEPT ERROR IN SCENARIO 1 OF RESNET50

| Class Name | Top MI (Concept) | | Lowest Distinctive (CRP) | | Lowest Distinctive (CRAFT) | | Concept Error |
|---|---|---|---|---|---|---|---|
| | 1st | 2nd | 1st | 2nd | 1st | 2nd | Value |
| person | car | backpack | car | backpack | car | tennis racket | 0.7291 |
| car | truck | bus | truck | traffic light | handbag | truck | 0.5385 |
| dining table | chair | cup | chair | fork | person | chair | 0.0166 |

From the information given in Table VI, it is evident that a poor-performing model is not ideal for concept-based explanations due to the lack of clear distinctions between classes. This can be seen in scenario 2 of ResNet50, where classes like person show less distinctiveness with other unrelated classes. In scenario 1, shown in Table V, we see a more effective distinction between highly confused classes like car and person, which indicates that a well-performing model actively tries to separate these difficult-to-distinguish classes (previously established based on confusion scores, see Table I, V, VI and III).

TABLE VI. MUTUAL INFORMATION, CONCEPT DISTINCTIVENESS, AND CONCEPT ERROR IN SCENARIO 2 OF RESNET50

| Class Name | Top MI (Concept) | | Lowest Distinctive (CRP) | | Lowest Distinctive (CRAFT) | | Concept Error |
|---|---|---|---|---|---|---|---|
| | 1st | 2nd | 1st | 2nd | 1st | 2nd | Value |
| person | car | tennis racket | backpack | bottle | backpack | umbrella | 0.8136 |
| car | truck | traffic light | bench | fire hydrant | backpack | boat | 0.6388 |
| dining table | cup | bottle | chair | fork | person | potted plant | 0.0753 |

By focusing on distinctiveness metrics and correlating them with confusion patterns in Table I and co-occurrence in Table III, we see that increasing concept distinctiveness can significantly aid in or point to improved model performance. This insight not only helps in diagnosing where models are struggling but also guides how to curate datasets and improve feature learning to reduce confusion and improve overall classification accuracy.

### C. *Environmental Concepts Reveal Dataset Biases*

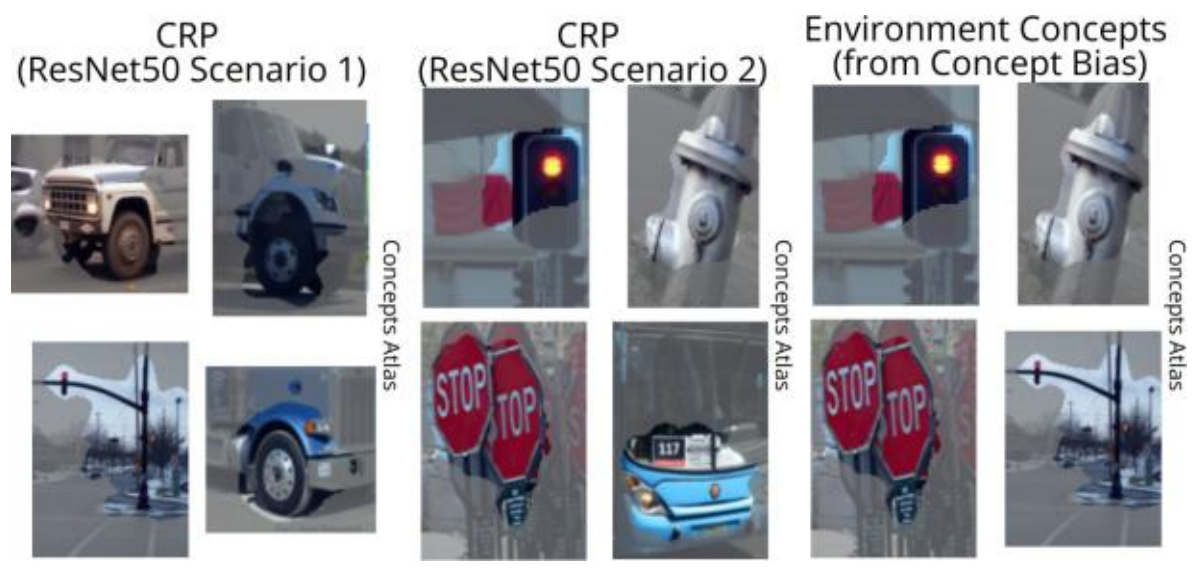

Figure 3. Environmental concepts generated from CRP for class “car” in scenario 1 and 2 of ResNet50.

TABLE VII. MUTUAL INFORMATION (CONCEPT), MUTUAL INFORMATION AND CONFUSION SCORES IN SCENARIO 1 OF VGG-16

| Class Name | Top MI (Concept) | | Top M (Class) | | Top Co fusion | |
|---|---|---|---|---|---|---|
| | 1st | 2nd | 1st | 2nd | 1st | 2nd |
| umbrella | person | handbag | backpack | handbag | person | car |
| dining table | chair | fork | chair | cup | apple | person |
| traffic light | person | car | car | fire hydrant | person | car |

Environmental concepts emerge from concept bias and often reflect patterns in the training dataset. We observe that classes within the same “supercategory” (e.g., *sports*: *baseball glove*, *tennis racket*) tend to produce biased explanations, frequently including environmental concepts from related classes, illustrated in Figure 3. This suggests that, beyond model performance, the diversity and distinctiveness of training samples play a key role in learning meaningful class representations.

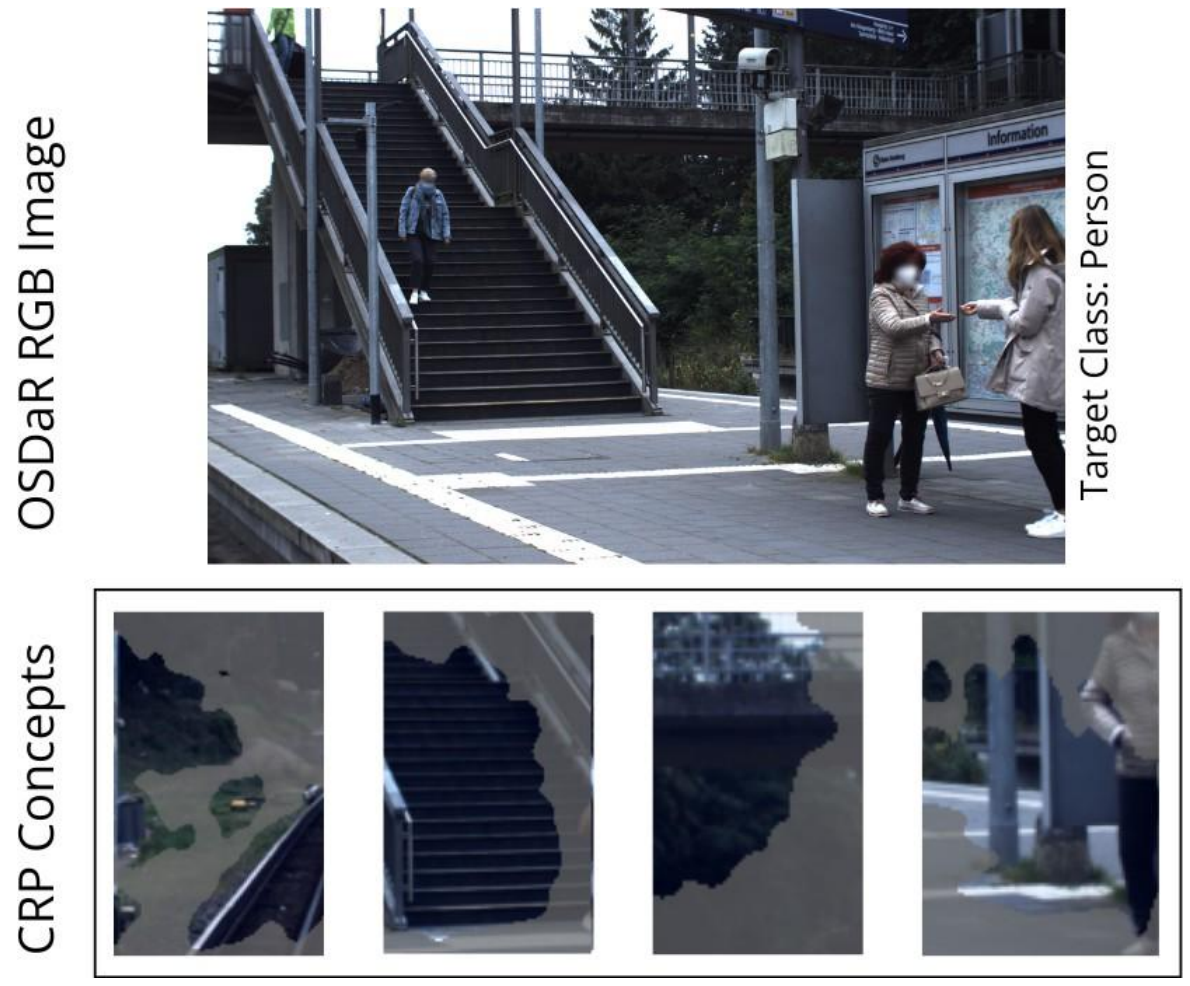

Figure 4. Concepts generated by CRP on OSDaR23 dataset for class "person".

Table VII illustrates the top mutual information and confusion scores for selected classes. For instance, *dining table* in scenario 1 is frequently associated with *chair*, *person*, *apple*, and *cup*, which are labels that share semantic but not structural similarity. Such associations, while intuitive to humans, suggest that the model is not generalizing but instead relying on frequent co-occurrences, which is problematic in deployed systems. High concept error rates for classes like *umbrella*, *person*, *handbag*, and *car*, paired with low distinctiveness scores between semantically unrelated objects (e.g., *umbrella* and *traffic light*), reinforce this concern, especially when models perform poorly.

To further support this, we evaluate OSDaR23 [25], a multi-sensor dataset for autonomous train driving. Despite strong accuracy (95.92%) and F1 (79.93) on a ResNet50 model trained on its RGB subset, CXAI explanations reveal low generalizability. Since *person* consistently appears near platforms or staircases, CRP visualizations heavily rely on these backgrounds, none of which are labeled in the dataset, as illustrated in Figure 4. As a result, *person* has the lowest distinctiveness score with *track*, and a high concept error, indicating dangerous misattribution.

These findings highlight how environmental concepts reveal dataset-induced biases that compromise generalization. In real-world or high-risk applications, such as autonomous systems, these misleading correlations can reduce model reliability. Diverse and well-annotated datasets are essential to prevent concept bias and ensure models learn robust, semantically accurate representations.

## V. Conclusion and Future Work

Our study demonstrates that confusion in multi-label classification is directly reflected in concept-based explanations. By comparing model evaluations with CXAI properties, we observe that label confusion often results from overlapping or spurious environmental concepts, emphasizing the role of CXAI in uncovering learning biases and assessing model generalizability. We further show that concept distinctiveness is inversely related to conceptual confusion, models with higher distinctiveness show clearer feature boundaries and reduced bias, while lower distinctiveness leads to shared or incorrect associations across classes. CRP and CRAFT help identify such conceptual ambiguities, making them useful tools for model diagnosis. Finally, our results highlight that environmental concepts can reveal dataset-induced biases, especially in cases where co-occurring objects affect model learning. In datasets with label imbalance or strong contextual patterns, models may form misleading correlations, reducing their ability to generalize. This is particularly problematic in high-risk applications, reinforcing the need for diverse, well-annotated datasets to ensure robust and reliable AI models. For future work, this case study can be extended to more complex models and datasets.

## Acknowledgement

This work is partly supported by BMFTR (Federal Ministry of Research, Technology and Space) in DAAD project 57616814 (SECAI, School of Embedded Composite AI, https://secai.org/).

## References

[1] Y. LeCun, Y. Bengio, and G. Hinton, “Deep learning,” *Nature*, vol. 521, pp. 436–444, 2015.

[2] W. Samek, G. Montavon, S. Lapuschkin, C. J. Anders, and K.-R. Müller, “Explaining deep neural networks and beyond: A review of methods and applications,” *Proceedings of the IEEE*, vol. 109, no. 3, pp. 247–278, 2021.

[3] E. Poeta, G. Ciravegna, E. Pastor, T. Cerquitelli, and E. Baralis, *Concept-based explainable artificial intelligence: A survey*, Preprint arXiv:2312.12936, 2023.

[4] R. Müller, M. Thoß, J. Ullrich, S. Seitz, and C. Knoll, “Interpretability is in the eye of the beholder: Human versus artificial classification of image segments generated by humans versus xai,” *International Journal of Human-Computer Interaction*, pp. 2371–2393, 2024.

[5] S. S. Y. Kim, E. A. Watkins, O. Russakovsky, R. Fong, and A. Monroy-Hernández, “"help me help the ai": Understanding how explainability can support human-ai interaction,” in *Proceedings of the 2023 CHI Conference on Human Factors in Computing Systems*, 2023, pp. 1–17.

[6] P. Stock and M. Cisse, “Convnets and imagenet beyond accuracy: Understanding mistakes and uncovering biases,” in *Proceedings of the European Conference on Computer Vision (ECCV)*, 2018, pp. 504–519.

[7] B. Kim et al., “Interpretability beyond feature attribution: Quantitative testing with concept activation vectors (tcav),” in *International Conference on Machine Learning*, PMLR, 2018, pp. 2668–2677.

[8] T.-Y. Lin et al., “Microsoft coco: Common objects in context,” in *European Conference on Computer Vision*, Springer, 2014, pp. 740–755.

[9] R. Achtibat et al., “From attribution maps to human-understandable explanations through concept relevance propagation,” *Nature Machine Intelligence*, vol. 5, pp. 1006–1019, 2023.

[10] T. Fel et al., “Craft: Concept recursive activation factorization for explainability,” in *Proceedings of the IEEE/CVF Conference on Computer Vision and Pattern Recognition*, 2023, pp. 2711–2721.

[11] J. H. Lee, G. Mikriukov, G. Schwalbe, S. Wermter, and D. Wolter, “Concept-based explanations in computer vision: Where are we and where could we go?” In *European Conference on Computer Vision*, Springer, 2025, pp. 266–287.

[12] K. K. Singh et al., “Don’t judge an object by its context: Learning to overcome contextual bias,” in *Proceedings of the IEEE/CVF Conference on Computer Vision and Pattern Recognition*, 2020, pp. 11 070–11 078.

[13] V. V. Ramaswamy, S. S. Y. Kim, R. Fong, and O. Russakovsky, “Overlooked factors in concept-based explanations: Dataset choice, concept learnability, and human capability,” in *Proceedings of the IEEE/CVF Conference on Computer Vision and Pattern Recognition*, 2023, pp. 10 932–10 941.

[14] A. Binder, G. Montavon, S. Lapuschkin, K.-R. Müller, and W. Samek, “Layer-wise relevance propagation for neural networks with local renormalization layers,” in *International Conference on Artificial Neural Networks*, Springer International Publishing, 2016, pp. 63–71.

[15] R. R. Selvaraju et al., “Grad-cam: Visual explanations from deep networks via gradient-based localization,” in *Proceedings of the IEEE international conference on computer vision*, 2017, pp. 618–626.

[16] K. He, X. Zhang, S. Ren, and J. Sun, “Deep residual learning for image recognition,” in *Proceedings of the IEEE Conference on Computer Vision and Pattern Recognition*, 2016, pp. 770–778.

[17] K. Simonyan and A. Zisserman, “Very deep convolutional networks for large-scale image recognition,” *arXiv preprint arXiv:1409.1556*, 2014.

[18] H. Ibrahim Aysel, X. Cai, and A. Prugel-Bennett, “Explainable artificial intelligence: Advancements and limitations,” *Applied Sciences*, vol. 15, p. 7261, 2025.

[19] J. Deng et al., “Imagenet: A large-scale hierarchical image database,” in *IEEE Conference on Computer Vision and Pattern Recognition*, IEEE, 2009, pp. 248–255.

[20] D. Krstinić, A. K. Skelin, I. Slapničar, and M. Braović, “Multi-label confusion tensor,” *IEEE Access*, vol. 12, pp. 9860–9870, 2024.

[21] B. Wang, L. Li, Y. Nakashima, and H. Nagahara, “Learning bottleneck concepts in image classification,” in *Proceedings of the IEEE/CVF Conference on Computer Vision and Pattern Recognition*, 2023, pp. 10 962–10 971.

[22] P. W. Koh et al., “Concept bottleneck models,” in *International Conference on Machine Learning*, PMLR, 2020, pp. 5338–5348.

[23] L. O’Mahony, V. Andrearczyk, H. Müller, and M. Graziani, “Disentangling neuron representations with concept vectors,” in *Proceedings of the IEEE/CVF Conference on Computer Vision and Pattern Recognition*, 2023, pp. 3770–3775.

[24] M. E. Zarlenga et al., *Concept embedding models: Beyond the accuracy-explainability trade-off*, Preprint arXiv:2209.09056, 2022.

[25] R. Tagiew et al., “Osdar23: Open sensor data for rail 2023,” in *International Conference on Robotics and Automation Engineering*, 2023, pp. 270–276.